\documentstyle[11pt]{article}

\title{\bf FASTUS:  A Cascaded Finite-State Transducer for Extracting 
	Information from Natural-Language Text}

\author{ Jerry R. Hobbs, Douglas Appelt, John Bear, \\ 
David Israel, Megumi Kameyama, \\ Mark Stickel, and Mabry Tyson \\
Artificial Intelligence Center \\
SRI International \\
Menlo Park, California}

\date{ }				

\begin{document}

\maketitle  

\begin{abstract}                 

FASTUS is a system for extracting information from natural language text
for entry into a database and for other applications.  It works
essentially as a cascaded, nondeterministic finite-state automaton.
There are five stages in the operation of FASTUS.  In Stage 1, names and
other fixed form expressions are recognized.  In Stage 2, basic noun
groups, verb groups, and prepositions and some other particles are
recognized.  In Stage 3, certain complex noun groups and verb groups are
constructed.  Patterns for events of interest are identified in Stage 4
and corresponding ``event structures'' are built.  In Stage 5, distinct
event structures that describe the same event are identified and merged,
and these are used in generating database entries.  This decomposition
of language processing enables the system to do exactly the right amount
of domain-independent syntax, so that domain-dependent semantic and
pragmatic processing can be applied to the right larger-scale
structures.  FASTUS is very efficient and effective, and has been used
successfully in a number of applications.

\end{abstract}

\section{Introduction}

{\bf FASTUS} is a (slightly permuted) acronym for Finite State Automaton
Text Understanding System.  It is a system for extracting information
from free text in English, Japanese, and potentially other languages as
well, for entry into a database and for other applications.  It works
essentially as a set of cascaded, nondeterministic finite-state
transducers.  Successive stages of processing are applied to the input,
patterns are matched, and corresponding composite structures are built.
The composite structures built in each stage provides the input to the
next stage.

In Section 2 we describe the information extraction task, especially as
exemplified by the Message Understanding Conference (MUC) evaluations
(Sundheim 1992, 1993), which originally motivated the system design.  We
also discuss the important distinction between information extraction
systems and text understanding systems.  Section 3 is a review of
previous finite-state approaches to natural language processing.
Section 4 describes the overall architecture of the FASTUS system, and
Sections 5 through 9 describe the individual stages.  Section 10
describes the history of the system, including its principal
applications and its performance in the MUC evaluations.  Section 11
summarizes the advantages of the FASTUS approach.

\section{The Information Extraction Task} 

There are a large number of applications in which a large corpus of
texts must be searched for particular kinds of information and that
information must be entered into a database for easier access.  In the
applications implemented so far, the corpora have typically been news
articles or telegraphic military messages.  The task of the system is to
build templates or database entries with information about who did what
to whom, when and where.

This task has been the basis of the successive MUC evaluations.  In
MUC-1 in June 1987, and MUC-2 in May 1989, the corpora were telegraphic
messages about naval operations.  The task definition for the
evaluations took shape over the course of these two efforts.

The corpus for MUC-3 in June 1991 and MUC-4 in June 1992 consisted of
news articles and transcripts of radio broadcasts, translated from
Spanish, from the Foreign Broadcast Information Service.  The focus of
the articles was Latin American terrorism.  The articles ranged from
one third of a page to two pages in length.  The template-filling task
required identifying, among other things, the perpetrators and victims
of each terrorist act described in an article, the occupations of the
victims, the type of physical entity attacked or destroyed, the date,
the location, and the effect on the targets.  Many articles described
multiple incidents, while other texts were completely irrelevant.

The following are some relevant excerpts from a sample terrorist report
(TST2-MUC4-0048).  
\begin{quotation}

   San Salvador, 19 Apr 89 (ACAN-EFE) -- [TEXT] Salvadoran
President-elect Alfredo Cristiani condemned the terrorist killing of
Attorney General Roberto Garcia Alvarado and accused the Farabundo
Marti National Liberation Front (FMLN) of the crime.

$\ldots$

   Garcia Alvarado, 56, was killed when a bomb placed by urban
guerrillas on his vehicle exploded as it came to a halt at an
intersection in downtown San Salvador.

$\ldots$

   Vice President-elect Francisco Merino said that when the attorney
general's car stopped at a light on a street in downtown San Salvador,
an individual placed a bomb on the roof of the armored vehicle.

$\ldots$

   According to the police and Garcia Alvarado's driver, who escaped
unscathed, the attorney general was traveling with two bodyguards.
One of them was injured.

\end{quotation}

Some of the corresponding database entries are as follows:
\begin{flushleft} \begin{tabular}{@{}p{1cm}ll}

&{\bf Incident: Date} &                   - 19 Apr 89 \\
&{\bf Incident: Location} &               El Salvador: San Salvador (city) \\
&{\bf Incident: Type} &                   Bombing \\
&{\bf Perpetrator: Individual ID} &       ``urban guerrillas'' \\
&{\bf Perpetrator: Organization ID} &     ``FMLN'' \\
&{\bf Perpetrator: Organization} &        Suspected or Accused by \\
& \hspace*{1cm} {\bf Confidence} &        \hspace*{1cm} Authorities:  ``FMLN'' \\
&{\bf Physical Target: Description} &     ``vehicle'' \\
&{\bf Physical Target: Effect} &          Some Damage:  ``vehicle'' \\
&{\bf Human Target: Name} &               ``Roberto Garcia Alvarado'' \\
&{\bf Human Target: Description} &        ``attorney general'': ``Roberto  \\
&  &					  \hspace*{1cm} Garcia Alvarado'' \\
&  & 					  ``driver'' \\
&  & 					  ``bodyguards'' \\
\end{tabular} \end{flushleft}
\begin{flushleft} \begin{tabular}{@{}p{1cm}ll}
&{\bf Human Target: Effect} &             Death: ``Roberto Garcia \\
&  &					  \hspace*{1cm} Alvarado'' \\
&  & 					  No Injury: ``driver'' \\
&  & 					  Injury: ``bodyguards'' \\

\end{tabular} \end{flushleft}

Fifteen sites participated in MUC-3, and seventeen in MUC-4.  A
development corpus of 1500 texts, together with their corresponding
templates and an automatic scoring program, were made available.  The
systems were tested on a new set of 100 messages from the same time
slice as the development messages, and in MUC-4 on a test set of 100
messages from a new time slice.

The task in MUC-5 in July 1993 was to extract information about 
joint ventures from business news, including the participants in the 
joint venture, the resulting company, the ownership and capitalization,
and the intended activity.

A typical text is the following:
\begin{quotation}

Bridgestone Sports Co. said Friday it has set up a joint venture
in Taiwan with a local concern and a Japanese trading house to produce
golf clubs to be shipped to Japan.

The joint venture, Bridgestone Sports Taiwan Co., capitalized at 20 million
new Taiwan dollars, will start production in January 1990 with production
of 20,000 iron and ``metal wood'' clubs a month.

\end{quotation}

This text is used as an example in the description below of the FASTUS
system.

The information to be extracted from this text is shown in the following
templates: 
\begin{flushleft}
\begin{tabular}{@{}p{5mm}p{45mm}p{70mm}@{}}
 & {\bf TIE-UP-1:} \\
 & Relationship: & TIE-UP \\
 & Entities: & ``Bridgestone Sports Co.'' \\
 & & ``a local concern'' \\
 & & ``a Japanese trading house'' \\
 & Joint Venture Company: & ``Bridgestone Sports Taiwan Co.'' \\
 & Activity: & ACTIVITY-1 \\
 & Amount: & NT\$20000000 \\

\end{tabular}
\end{flushleft} \vspace{-2mm}

\begin{flushleft}
\begin{tabular}{@{}p{5mm}p{45mm}p{70mm}@{}}

 & {\bf ACTIVITY-1:} \\
 & Activity: & PRODUCTION \\
 & Company: & ``Bridgestone Sports Taiwan Co.'' \\
 & Product: & ``iron and `metal wood' clubs''\\
 & Start Date: & DURING: January 1990 \\

\end{tabular}
\end{flushleft} \vspace{-2mm}

Seventeen sites participated in MUC-5.  It was conducted in conjunction
with the ARPA-sponsored Tipster program, whose objective has been to
encourage development of information extraction technology and to move
it into the user community.

The principal measures for information extraction tasks are recall and
precision.  {\it Recall} is the number of answers the system got right
divided by the number of possible right answers.  It measures how
complete or comprehensive the system is in its extraction of relevant
information.  {\it Precision} is the number of answers the system got
right divided by the number of answers the system gave.  It measures the
system's correctness or accuracy.  For example, if there are 100
possible answers and the system gives 80 answers and gets 60 of them
right, its recall is 60\% and its precision is 75\%.

In addition, a combined measure, called the F-score, is often used.  It
is an approximation to the weighted geometric mean of recall and
precision.  The F-score is defined as follows:  
\begin{center}

	$F = \frac{(\beta^{2} + 1) P R}{\beta^{2} P + R}$

\end{center}
where $P$ is precision, $R$ is recall, and $\beta$ is a parameter
encoding the relative importance of recall and precision.  If $\beta =
1$, they are weighted equally.  If $\beta > 1$, precision is more
significant; if $\beta < 1$, recall is.   

It is important to distinguish between two types of natural language
systems:  {\it information extraction} systems and {\it text
understanding} systems.  In information extraction, 
\begin{itemize}

\item generally only a fraction of the text is relevant; for example, in
the case of the MUC-4 terrorist reports, probably only about 10\% of the
text was relevant;

\item information is mapped into a predefined, relatively simple, rigid
target representation; this condition holds whenever entry of
information into a database is the task;

\item the subtle nuances of meaning and the writer's goals in writing
the text are of at best secondary interest.

\end{itemize}

This contrasts with text understanding, where 
\begin{itemize}

\item the aim is to make sense of the entire text;

\item the target representation must accommodate the full complexities
of language;

\item one wants to recognize the nuances of meaning and the writer's
goals.

\end{itemize}

The task in the MUC evaluations has been information extraction, not 
text understanding.  When SRI participated in the MUC-3 evaluation
in 1991, we used TACITUS, a text-understanding system (Hobbs et al.,
1992; Hobbs et al., 1993).  Using it for the information extraction
task gave us a high precision, the highest of any of the sites.
However, because it was spending so much of its time attempting to make
sense of portions of the text that were irrelevant to the task, the
system was extremely slow.  As a result, development time was slow,
and consequently recall was mediocre.

FASTUS, by contrast, is an information extraction system, rather than a
text understanding system.  Our original motivation in developing FASTUS
was to build a system that was more appropriate to the information
extraction task.

Although information extraction is not the same as full text
understanding, there are many important applications for information
extraction systems, and the technology promises to be among the first
genuinely practical applications of natural language processing.

\section{The Finite-State Approach}

The inspiration for FASTUS was threefold.  First, we were struck by the
strong performance in MUC-3 that the group at the University of
Massachusetts got out of a fairly simple system (Lehnert et al., 1991).
It was clear they were not doing anything like the depth of
preprocessing, syntactic analysis, or pragmatics that was being done by
the systems at SRI, General Electric, or New York University.  They were
not doing a lot of processing.  But they were doing the {\it right}
processing for the task. 

The second source of inspiration was Pereira's work on finite-state
approximations of grammars (Pereira, 1990).  We were especially
impressed by the speed of the implemented system.

Our desire for speed was the third impetus for the development of
FASTUS.  It was simply too embarassing to have to report at the MUC-3
conference that it took TACITUS 36 hours to process 100 messages.
FASTUS brought that time down to less than 12 minutes.

Finite-state models are clearly not adequate for full natural language
processing.  However, if context-free parsing is not cost-effective when
applied to real-world text, then an efficient text processor might make
use of weaker language models, such as regular or finite-state grammars.
Every computational linguistics graduate student knows, from the first
textbook that introduces the Chomsky hierarchy, that English has
constructs, such as center embedding, that cannot be described by any
finite-state grammar.  This fact biased researchers away from serious
consideration of possible applications of finite-state grammars to
difficult problems.

Church (1980) was the first to advocate finite-state grammars as a
processing model for language understanding.  He contended that,
although English is clearly not a regular language, memory limitations
make it impossible for people to exploit that context-freeness in its
full generality, and therefore a finite-state mechanism might be
adequate in practice as a model of human linguistic performance.  A
computational realization of memory limitation as a depth cutoff was
implemented by Black (1989).

Pereira and Wright (1991) developed methods for constructing
finite-state grammars from context free grammars that overgenerate in
certain systematic ways.  The finite-state grammar could be applied in
situations, for example, as the language model in a speech understanding
system, where computational considerations are paramount.

At this point, the limitations of the application of finite-state
grammars to natural-language processing have not yet been determined.
We believe our research has established that these simple mechanisms can
achieve a lot more than had previously been thought possible.

\section{Overview of the FASTUS Architecture}

The key idea in FASTUS, the ``cascade'' in ``cascaded
finite-state automata'', is to separate processing into several stages.
The earlier stages recognize smaller linguistic objects and work in a
largely domain-independent fashion.  They use purely linguistic
knowledge to recognize that portion of the syntactic structure of the
sentence that linguistic methods can determine reliably, requiring
little or no modification or augmentation as the system is moved from
domain to domain.  These stages have been implemented for both English
and Japanese.

The later stages take these linguistic objects as input and find
domain-dependent patterns among them.

The current version of FASTUS may be thought of as using five levels of
processing:  
\begin{enumerate}

\item Complex Words:  This includes the recognition of multiwords and
proper names.

\item Basic Phrases:  Sentences are segmented into noun groups,
verb groups, and particles.

\item Complex Phrases:  Complex noun groups and complex verb groups
are identified.

\item Domain Events:  The sequence of phrases produced at Level 3 is
scanned for patterns for events of interest to the application, and when
they are found, structures are built that encode the information about
entities and events contained in the pattern.

\item Merging Structures:  Structures arising from different parts of
the text are merged if they provide information about the same entity or
event.

\end{enumerate}

As we progress through the five levels, larger segments of text are
analyzed and structured.

This decomposition of the natural-language problem into levels is
essential to the approach.  Many systems have been built to do pattern
matching on strings of words.  One of the crucial innovations in our
approach has been dividing that process into separate levels for
recognizing phrases and recognizing event patterns.  Phrases can be
recognized reliably with purely syntactic information, and they provide
precisely the elements that are required for stating the event patterns
of interest.

Various versions of the system have had other, generally preliminary
stages of processing.  For the MUC-4 system we experimented with
spelling correction.  The experiments indicated that spelling
correction hurt, primarily because novel proper names got corrected to
other words, and hence were lost.

The MUC-4 system also had a preliminary stage in which each sentence was
first searched for trigger words.  At least one, generally low-frequency
trigger word was included for each pattern of interest that had been
defined.  For example, in the pattern 
\begin{verse}

	take $<$HumanTarget$>$ hostage

\end{verse} 
``hostage'' rather than ``take'' is the trigger word.  Triggering
reduced the processing time by about a third, but since it is hard to
maintain in a way that does not reduce recall and since the system is so
fast anyway, this stage has not been a part of subsequent versions of
the system.

We currently have a version of the system, a component in the Warbreaker
Message Handler System, for handling military messages about
time-critical targets, which has a preliminary stage of processing that
identifies the free and formatted portions of the messages, breaks the
free text into sentences, and identifies tables, outlines, and lists.
The table processing is described in Tyson et al. (to appear).

At one point we investigated incorporating a part-of-speech tagger into
the system.  This turned out to double the run-time of the entire
system, and it made similar mistakes to those that the basic phrase
recognition stage made.  Consequently, we have not used this component.

Every version of the system we have built has included a postprocessing 
stage that converts the event structures into the format required by
the application or evaluation.

The system is implemented in CommonLisp and runs on Sun workstations.
Several partial implementations of FASTUS in C++ have been built.

\section{Complex Words} 

The first level of processing identifies multiwords such as ``set up'',
``trading house'', ``new Taiwan dollars'', and ``joint venture'', and
company names like ``Bridgestone Sports Co.'' and ``Bridgestone Sports
Taiwan Co.''.  The names of people and locations, dates, times, and
other basic entities are also recognized at this level.

Languages in general are very productive in the construction of short,
multiword fixed phrases and proper names employing specialized
microgrammars, and this is the level at which they are recognized.

Not all names can be recognized by their internal structure.  Thus,
there are rules in subsequent stages for recognizing unknown possible
names as names of specific types.  For example, in 
\begin{verse}

	XYZ's sales \\
	Vaclav Havel, 53, president of the Czech Republic,

\end{verse}
we might not know that XYZ is a company and Vaclav Havel is a person,
but the immediate context establishes that.

\section{Basic Phrases}

The problem of syntactic ambiguity is AI-complete.  That is, we will not
have systems that reliably parse English sentences correctly until we
have encoded much of the real-world knowledge that people bring to bear
in their language comprehension.  For example, noun phrases cannot be
reliably identified because of the prepositional phrase attachment
problem.  However, certain syntactic constructs can be reliably
identified.  One of these is the noun group, that is, the head noun of a
noun phrase together with its determiners and other left modifiers.
Another is what we are calling the ``verb group'', that is, the verb
together with its auxiliaries and any intervening adverbs.  Moreover,
an analysis that identifies these elements gives us exactly the units we
most need for domain-dependent processing.

Stage 2 in FASTUS identifies noun groups, verb groups, and several
critical word classes, including prepositions, conjunctions, relative
pronouns, and the words ``ago'' and ``that''.  Phrases that are
subsumed by larger phrases are discarded.  Pairs of overlapping,
nonsubsuming phrases are rare, but where they occur both phrases are
kept.  This sometimes compensates for an incorrect analysis in Stage 2.

The first sentence in the sample joint venture text is segmented by
Stage 2 into the following phrases:  
\begin{flushleft}
\begin{tabular}{@{}p{25mm}p{35mm}p{80mm}@{}}

 & Company Name: & Bridgestone Sports Co. \\
 & Verb Group: & said \\
 & Noun Group: & Friday \\
 & Noun Group: & it \\
 & Verb Group: & had set up \\
 & Noun Group: & a joint venture \\
 & Preposition: & in \\
 & Location: & Taiwan \\
 & Preposition: & with \\
 & Noun Group: & a local concern \\
 & Conjunction: & and \\
 & Noun Group: & a Japanese trading house \\
 & Verb Group: & to produce \\
 & Noun Group: & golf clubs \\
 & Verb Group: & to be shipped \\
\end{tabular}
\begin{tabular}{@{}p{25mm}p{35mm}p{80mm}@{}}
 & Preposition: & to \\
 & Location: & Japan \\

\end{tabular}
\end{flushleft} \vspace{-2mm}

``Company Name'' and ``Location'' are special kinds of noun group.

Noun groups are recognized by a finite-state grammar that 
encompasses most of the complexity that can occur in
English noun groups, including numbers, numerical modifiers like
``approximately'', other quantifiers and determiners, participles in
adjectival position, comparative and superlative adjectives, conjoined
adjectives, and arbitrary orderings and conjunctions of prenominal nouns
and noun-like adjectives.  Thus, among the noun groups recognized are
\begin{verse}

	approximately 5 kg \\
	more than 30 people \\
	the newly elected president \\
	the largest leftist political force \\
	a government and commercial project 

\end{verse} 

Verb groups are recognized by a finite-state grammar that tags them as
Active, Passive, Gerund, and Infinitive.  Verbs are sometimes locally
ambiguous between active and passive senses, as the verb ``kidnapped''
in the two sentences, 
\begin{verse}

	Several men kidnapped the mayor today. \\
	Several men kidnapped yesterday were released today.

\end{verse} 
These are tagged as Active/Passive, and Stage 4 resolves the
ambiguity if necessary.

Predicate adjective constructions are also recognized and classified as
verb groups.

The grammars for noun groups and verb groups used in MUC-4 are given in
Hobbs et al. (1992); although these grammars have subsequently been 
augmented for domain-specific constructs, the core remains essentially 
the same.

Unknown or otherwise unanalyzed words are ignored in subsequent
processing, unless they occur in a context that indicate they could be
names.

The breakdown of phrases into nominals, verbals, and particles is a
linguistic universal.  Whereas the precise parts of speech that occur in
any language can vary widely, every language has elements that are
fundamentally nominal in character, elements that are fundamentally
verbal or predicative, and particles or inflectional affixes that encode
relations among the other elements (Croft, 1991).

\section{Complex Phrases} 

In Stage 3, complex noun groups and verb groups that can be recognized
reliably on the basis of domain-independent, syntactic information are
recognized.  This includes the attachment of appositives to their head
noun group, 
\begin{verse}

	The joint venture, Bridgestone Sports Taiwan Co.,

\end{verse}
the construction of measure phrases,
\begin{verse}

	20,000 iron and ``metal wood'' clubs a month,

\end{verse}
and the attachment of ``of'' and ``for'' prepositional phrases to
their head noun groups,
\begin{verse}

	production of 20,000 iron and ``metal wood'' clubs a month.

\end{verse}
Noun group conjunction, 
\begin{verse}

	a local concern and a Japanese trading house, 

\end{verse}
is done at this level as well. 

In the course of recognizing basic and complex phrases, entities and
events of domain interest are often recognized, and the structures for
these are constructed.  In the sample joint-venture text, entity
structures are constructed for the companies referred to by the phrases
``Bridgestone Sports Co.'', ``a local concern'', ``a Japanese trading
house'', and ``Bridgestone Sports Taiwan Co.''  Information about
nationality derived from the words ``local'' and ``Japanese'' is
recorded.  Corresponding to the complex noun group ``The joint venture,
Bridgestone Sports Taiwan Co.,'' the following relationship structure is
built:  
\begin{flushleft} 
\begin{tabular}{@{}p{5mm}p{45mm}p{70mm}@{}}

 & Relationship: & TIE-UP \\
 & Entities: & -- \\
 & Joint Venture Company: & ``Bridgestone Sports Taiwan Co.'' \\
 & Activity: & -- \\
 & Amount: & -- \\

\end{tabular}
\end{flushleft} \vspace{-2mm}
Corresponding to the complex noun group ``production
of 20,000 iron and `metal wood' clubs a month'', the following 
activity structure is built up:
\begin{flushleft}
\begin{tabular}{@{}p{5mm}p{45mm}p{70mm}@{}}

 & Activity: & PRODUCTION \\
 & Company: & -- \\
 & Product: & ``iron and `metal wood' clubs''\\
 & Start Date: & -- \\

\end{tabular}
\end{flushleft} \vspace{-2mm}

When we first implemented the Complex Phrase level of processing, our
intention was to use it only for complex noun groups, as in the
attachment of ``of'' prepositional phrases to head nouns.  Then in the
final week before an evaluation, we wanted to make a change in what
sorts of verbs were accepted by a set of patterns; this change, though,
would have required our making extensive changes in the domain patterns.
Rather than do this at such a late date, we realized it would be easier
to define a complex verb group at the Complex Phrase level.  We then
immediately recognized that this was not an {\it ad hoc} device, but in
fact the way we should have been doing things all along.  We had
stumbled onto an important property of language--complex verb
groups---whose exploitation would have resulted in a significant
simplification in the rules for the Stage 4 patterns.

Consider the following variations:
\begin{verse}

	GM {\it formed} a joint venture with Toyota. \\
	GM {\it announced it was forming} a joint venture with Toyota. \\
	GM {\it signed an agreement forming} a joint venture with Toyota. \\
	GM {\it announced it was signing an agreement to form} a joint venture
		with Toyota. 

\end{verse}
Although these sentences may differ in significance for some
applications, they were equivalent in meaning within the MUC-5
application and would be in many others.  Rather than defining each of
these variations, with all their syntactic variants, at the domain
pattern level, the user should be able to define complex verb groups
that share the same significance.  Thus, ``formed'', ``announced it was
forming'', ``signed an agreement forming'', and ``announced it was
signing an agreement to form'' are all equivalent, at least in this
application, and once they are defined to be so, only one Stage 4
pattern needs to be expressed.

Various modalities can be associated with verb groups.  In
\begin{verse}

	GM will form a joint venture with Toyota.

\end{verse}
the status of the joint venture is ``Planned'' rather than ``Existing''.
But the same is true in the following sentences.
\begin{verse}

	GM plans to form a joint venture with Toyota. \\
	GM expects to form a joint venture with Toyota. \\
	GM announced plans to form a joint venture with Toyota. 

\end{verse}
Consequently, as patterns are defined for each of these complex verb
groups, the correct modality can be associated with them as well.

Verb group conjunction, as in
\begin{verse}

	Terrorists {\it kidnapped and killed} three people.

\end{verse}
is handled at this level as well.  

Our current view is that this stage of processing corresponds to an
important property of human languages.  In many languages some adjuncts
are more tightly bound to their head nouns than others.  ``Of''
prepositional phrases are in this category, as are phrases headed by
prepositions that the head noun subcategorizes for.  The basic noun
group together with these adjuncts constitutes the complex noun group.
Complex verb groups are also motivated by considerations of linguistic
universality.  Many languages have quite elaborate mechanisms for
constructing complex verbs.  One example in English is the use of
control verbs; ``to conduct an attack'' means the same as ``to attack''.
Many of these higher operators shade the core meaning with a
modality, as in ``plan to attack'' and ``fail to attack''.

\section{Domain Events}

The input to Stage 4 of FASTUS is a list of complex phrases in the order
in which they occur.  Anything that is not included in a basic or
complex phrase in Stage 3 is ignored in Stage 4; this is a significant
source of the robustness of the system.  Patterns for events of interest
are encoded as finite-state machines, where state transitions are
effected by phrases.  The state transitions are driven off the head
words in the phrases.  That is, each pair of relevant head word and
phrase type---such as ``company-NounGroup'',
``formed-PassiveVerbGroup'', ``bargaining-NounGroup'', and
``bargaining-PresentParticipleVerbGroup''--- has an associated set of
state transitions.

In the sample joint-venture text, the domain event patterns 
\begin{verse}

	$<$Company/ies$>$ $<$Set-up$>$ $<$Joint-Venture$>$ with
		$<$Company/ies$>$ 

\end{verse}
and
\begin{verse}

	$<$Produce$>$ $<$Product$>$ 

\end{verse} 
are instantiated in the first sentence, and the patterns 
\begin{verse}

	$<$Company$>$ $<$Capitalized$>$ at $<$Currency$>$ 

\end{verse}
and
\begin{verse}

	$<$Company$>$ $<$Start$>$ $<$Activity$>$ in/on $<$Date$>$ 

\end{verse} 
are instantiated in the second.  These four patterns result in the
following four structures being built:
\begin{flushleft}
\begin{tabular}{@{}p{5mm}p{45mm}p{70mm}@{}}

 & Relationship: & TIE-UP \\
 & Entities: & ``Bridgestone Sports Co.'' \\
 & & ``a local concern'' \\
 & & ``a Japanese trading house'' \\
 & Joint Venture Company: & -- \\
 & Activity: & -- \\
 & Amount: & -- \\

\end{tabular}
\end{flushleft} \vspace{-2mm}

\begin{flushleft}
\begin{tabular}{@{}p{5mm}p{45mm}p{70mm}@{}}

 & Activity: & PRODUCTION \\
 & Company: & -- \\
 & Product: & ``golf clubs''\\
 & Start Date: & -- \\

\end{tabular}
\end{flushleft} \vspace{-2mm}

\begin{flushleft}
\begin{tabular}{@{}p{5mm}p{45mm}p{70mm}@{}}

 & Relationship: & TIE-UP \\
 & Entities: & -- \\
 & Joint Venture Company: & ``Bridgestone Sports Taiwan Co.'' \\
 & Activity: & -- \\
 & Amount: & NT\$20000000 \\

\end{tabular}
\end{flushleft} \vspace{-2mm}
(This is an augmentation of the previous relationship structure.)

\begin{flushleft}
\begin{tabular}{@{}p{5mm}p{45mm}p{70mm}@{}}

 & Activity: & PRODUCTION \\
 & Company: & ``Bridgestone Sports Taiwan Co.'' \\
 & Product: & -- \\
 & Start Date: & DURING: January 1990 \\

\end{tabular}
\end{flushleft} \vspace{-2mm}

Although subjects are always obligatory in main clauses, it was
determined in the MUC-4 evaluation that better performance in both
recall and precision were obtained if the system generated an event
structure from a verb together with its object, even if its subject
could not be determined.

A certain amount of ``pseudo-syntax'' is done in Stage 4.  The material
between the end of the subject noun group and the beginning of the main
verb group must be read over.  There are patterns to accomplish this.
Two of them are as follows:  
\begin{verse}

	Subject \{Preposition NounGroup\}* VerbGroup

	Subject Relpro \{NounGroup $\mid$ Other\}* VerbGroup \{NounGroup $\mid$ Other\}* VerbGroup

\end{verse} 
The first of these patterns reads over prepositional phrases.  The
second over relative clauses.  The verb group at the end of these
patterns takes the subject noun group as its subject.  There is another
set of patterns for capturing the content encoded in relative clauses,
of the form 
\begin{verse}

	Subject Relpro \{NounGroup $\mid$ Other\}* VerbGroup

\end{verse} 
The finite-state mechanism is nondeterministic.  With the exception of
passive clauses subsumed by larger active clauses, all events that are
discovered in this stage of processing are retained.  Thus, the full
content can be extracted from the sentence 
\begin{verse}

	The mayor, who was kidnapped yesterday, was found dead today.

\end{verse} 
One branch discovers the incident encoded in the relative clause.
Another branch marks time through the relative clause and then discovers
the incident in the main clause.  These incidents are then merged.

A similar device is used for conjoined verb phrases.  The pattern
\begin{verse}

	Subject VerbGroup \{NounGroup $\mid$ Other\}* Conjunction VerbGroup

\end{verse} 
allows the machine to nondeterministically skip over the first conjunct
and associate the subject with the verb group in the second conjunct.
That is, when the first verb group is encountered, all its complements
and adjuncts are skipped over until a conjunction is encountered, and
then the subject is associated with a verb group, if that is what comes
next.  Thus, in the sentence 
\begin{verse}

	Salvadoran President-elect Alfredo Cristiani condemned the
		terrorist killing of Attorney General Roberto Garcia 
		Alvarado and accused the Farabundo Marti National 
		Liberation Front (FMLN) of the crime.

\end{verse} 
one branch will recognize the killing of Garcia and another the fact
that Cristiani accused the FMLN.

In addition, irrelevant event adjuncts in the verb phrase are read over
while relevant adjuncts are being sought.

Many subject-verb-object patterns are of course related to each other.
The sentence 
\begin{verse}

	GM manufactures cars.

\end{verse}
illustrates a general pattern for recognizing a company's activities.
But the same semantic content can appear in a variety of ways, including
\begin{verse}

	Cars are manufactured by GM. \\
	$\ldots$ GM, which manufactures cars$\ldots$ \\
	$\ldots$ cars, which are manufactured by GM$\ldots$  \\
	$\ldots$ cars manufactured by GM $\ldots$  \\
	GM is to manufacture cars. \\
	Cars are to be manufactured by GM. \\
	GM is a car manufacturer. 

\end{verse}
These are all systematically related to the active form of the sentence.
Therefore, there is no reason a user should have to specify all the
variations.  The FASTUS system is able to generate all of the variants
of the pattern from the simple active (S-V-O) form.

These transformations are executed at compile time, producing the more
detailed set of patterns, so that at run time there is no loss of
efficiency.

Various sorts of adjuncts can appear at virtually any place in these
patterns:
\begin{verse}

	Cars were manufactured last year by GM. \\
	Cars are manufactured in Michigan by GM. \\
	The cars, a spokesman announced, will be manufactured in California
		and Tennessee by General Motors.

\end{verse}
Again, these possibilities are systematic and predictable, so there is
no reason that the user should be burdened with defining separate
patterns for them.  Adjuncts are thus added automatically to patterns,
and the information, say, about date and place, is extracted from them.

In this way, the user, simply by observing and stating that a particular
S-V-O triple conveys certain items of information, is able to define
dozens of patterns in the run-time system.

This feature is not merely a clever idea for making a system more
convenient.  It rests on the fundamental idea that underlies generative
transformational grammar, but is realized in a way that does not impact
the efficiency of processing.  

The Stage 4 level of processing corresponds to the basic clause level
that characterizes all languages, the level at which in English
Subject-Verb-Object (S-V-O) triples occur, and thus again corresponds to
a linguistic universal.  This is the level at which predicate-argument
relations between verbal and nominal elements are expressed in their
most basic form.

\section{Merging Structures} 

The first four stages of processing all operate within the bounds of
single sentences.  The final level of processing operates over the whole
text.  Its task is to see that all the information collected about
a single entity or relationship is combined into a unified whole.  This
is one of the primary ways the problem of coreference is dealt with in
our approach.

The three criteria that are taken into account in determining whether
two structures can be merged are the internal structure of the noun
groups, nearness along some metric, and the consistency, or more
generally, the compatibility of the two structures.  

In the analysis of the sample joint-venture text, we have produced three
activity structures.  They are all consistent because they are all of
type PRODUCTION and because ``iron and `metal wood' clubs'' is
consistent with ``golf clubs''.  Hence, they are merged, yielding
\begin{flushleft} 
\begin{tabular}{@{}p{5mm}p{45mm}p{70mm}@{}}

 & Activity: & PRODUCTION \\
 & Company: & ``Bridgestone Sports Taiwan Co.'' \\
 & Product: & ``iron and `metal wood' clubs''\\
 & Start Date: & DURING: January 1990 \\

\end{tabular}
\end{flushleft} \vspace{-2mm}

Similarly, the two relationship structures that have been generated are
consistent with each other, so they are merged, yielding,
\begin{flushleft}
\begin{tabular}{@{}p{5mm}p{45mm}p{70mm}@{}}

 & Relationship: & TIE-UP \\
 & Entities: & ``Bridgestone Sports Co.'' \\
 & & ``a local concern'' \\
 & & ``a Japanese trading house'' \\
 & Joint Venture Company: & ``Bridgestone Sports Taiwan Co.'' \\
 & Activity: & -- \\
 & Amount: & NT\$20000000 \\

\end{tabular}
\end{flushleft} \vspace{-2mm}

Both of these cases are examples of identity coreference, where the
activities or relationships are taken to be identical.  We also handle
examples of inferential coreference here.  A joint venture has been
mentioned, a joint venture implies the existence of an activity, and an
activity has been mentioned.  It is consistent to suppose the activity
mentioned is the same as the activity implied, so we do.  The Activity
field of the Tie-Up structure is filled with a pointer to the Activity
structure.

For a given domain, there can be fairly elaborate rules for determining
whether two noun groups corefer, and thus whether their corresponding
entity structures should be merged.  A name can corefer with a
description, as ``General Motors'' with ``the company'', provided the
description is consistent with the other descriptions for that name.  A
precise description, like ``automaker'', can corefer with a vague
description, such as ``company'', with the precise description as the
result.  Two precise descriptions can corefer if they are semantically
compatible, like ``automaker'' and ``car manufacturer''.  In MUC-4 it
was determined that if two event structures had entities with proper
names in some of the role slots, they should be merged only if there was
an overlap in the names.

\section{History of the FASTUS System}

FASTUS was originally conceived, in December 1991, as a preprocessor for
TACITUS that could also be run in a stand-alone mode.  It was only in
the middle of May 1992, considerably later in our development, that we
decided the performance of FASTUS on the MUC-4 task was so high that we
could make FASTUS our complete system.

Most of the design work for the FASTUS system took place during January
1992.  The ideas were tested out on finding incident locations and
proper names in February.  With some initial favorable results in hand,
we proceeded with the implementation of the system in March.  The
implementation of Stages 2 and 3 was completed in March, and the general
mechanism for Stage 4 was completed by the end of April.  On May 6, we
did the first test of the FASTUS system on a blind test set of 100
terrorist reports, which had been withheld as a fair test, and we
obtained a score of 8\%\ recall and 42\%\ precision.  At that point we
began a fairly intensive effort to hill-climb on all 1300 development
texts then available, doing periodic runs on the fair test to monitor
our progress. This effort culminated in a score of 44\%\ recall and
57\%\ precision in the wee hours of June 1, when we decided to run the
official test.  The rate of progress was rapid enough that even a few
hours of work could be shown to have a noticeable impact on the score.
Our scarcest resource was time, and our supply of it was eventually
exhausted well before the point of diminishing returns.

We were thus able, in three and a half weeks, to increase the system's
F-score by 36.2 points, from 13.5 to 49.7.

In the actual MUC-4 evaluation, on a blind test of 100 texts, we
achieved a recall of 44\%\ with precision of 55\%\ using the most
rigorous penalties for missing and spurious fills.  This corresponds to
an F-score ($\beta = 1$) of 48.9.  On the second blind test of 100
texts, covering incidents from a different time span than the training
data, we observed, surprisingly, an identical recall score of 44\%;
however our precision fell to 52\%, for an F-score of 47.7.  It was
reassuring to see that there was very little degradation in performance
when moving to a time period over which the system had not been trained.

Out of the seventeen sites participating in MUC-4, only General
Electric's system performed significantly better (a recall of 62\% and a
precision of 53\% on the first test set), and their system had been
under development for over five years (Sundheim, 1992).  Given our
experience in bringing the system to its current level of performance in
three and a half weeks, we felt we could achieve results in that range
with another month or two of effort.  Studies indicate that human
intercoder reliability on information extraction tasks is in the 65-80\%
range.  Thus, we believe this technology can perform at least 75\% as
well as humans.

And considerably faster.  One entire test set of 100 messages, ranging
from a third of a page to two pages in length, required 11.8 minutes of
CPU time on a Sun SPARC-2 processor.  The elapsed real time was 15.9
minutes, although observed time depends on the particular hardware
configuration involved.

In more concrete terms, this means that FASTUS could read 2,375 words per
minute.  It could analyze one text in an average of 9.6 seconds.  This
translates into 9,000 texts per day.

The FASTUS system was an order of magnitude faster than the other
leading systems at MUC-4.

This fast run time translates directly into fast development time, and
was the reason we could improve the scores so rapidly during May 1992.

A new version of the FASTUS system was developed in the following year,
and it was used for the MUC-5 evaluation.  The most significant addition
was a convenient graphical user interface for defining rules, utilizing
SRI's Grasper system (Karp et al., 1993).  This made it much easier to
specify the state transitions of the finite-state machines defined for
the domain application.  In addition, it was at this point that Stages
2 and 3 were made separate stages of processing.

SRI entered the Japanese task in MUC-5 as well as the English.  We had
during the year developed a Japanese version of FASTUS for use in a
conference room reservation task for a commercial client.  This system
read and extracted the relevant information from romanji input, and was
later developed into a real-time spontaneous dialogue summarizer
(Kameyama et al., 1995).  For MUC-5 we converted this to handle kanji
characters as well, and used the Grasper-based interface to define rules
for recognizing joint ventures in both English and Japanese business
news.

In the English portion of the evaluation, FASTUS achieved a recall of
34\% and a precision of 56\%, for an F-score ($\beta = 1$) of 42.67.  In
the Japanese task, the system achieved a recall of 34\% and a precision
of 62\%, for an F-score of 44.21.  Four of the sites were part of the 
Tipster program, and as such received funding and several extra months
to work on the domain; SRI was not at that point in the Tipster program.
FASTUS outperformed all of the other non-Tipster systems.  Of the four
Tipster systems, only two outperformed FASTUS, and only one significantly
so.

In early 1994 we developed a declarative specification language called
FastSpec.  If the Grasper-based specification interface is like
augmented transition networks, then FastSpec is like unification
grammar.  The patterns are specified by regular grammars, the
applicability of the rules is conditioned on attributes associated with
the terminal symbols, and attributes can be set on the objects
constructed.

This new version of FASTUS has been used for a number of applications.
For one commercial client, we helped in the conversion of parts of the
FASTUS system to C++ for the purposes of name recognition.  For another
commercial client, a pilot version of FASTUS was included in a document
analysis system to aid researchers in discovering the ontology underlying
complex Congressional bills, thereby ensuring the consistency of laws
with the regulations that implement them.

In collaboration with E-Systems, SRI has developed the Warbreaker
Message Handling System, for extracting information about time-critical
targets from a large variety of military messages.  This incorporates
FASTUS as the component for handling the free text portions of the
messages.

For the dry run of the MUC-6 evaluation in April 1995, we implemented a
set of FastSpec rules for recognizing information about labor
negotiations, their participants, and the status of the talks.

SRI has also been involved in the second phase of the Tipster program.
As part of this effort, we have made FASTUS compliant with the Tipster
architecture, aimed at enabling several different document detection
and information extraction systems to interact as components in a 
single larger system.  

The successive versions of FASTUS represent steps toward making it more
possible for the nonexpert user to define his or her own patterns.  This
effort is continuing in our current projects.

\section{Conclusions}

Finite-state technology is sometimes characterized as {\it ad hoc}
and as {\it mere} pattern-matching.  However, our approach of using a
{\it cascade} of finite-state machines, where each level corresponds to
a linguistic natural kind, reflects important universals about language.
It was inspired by the remarkable fact that very diverse languages all
show the same nominal element - verbal element - particle distinction
and the basic phrase - complex phrase distinction.  Organizing a system
in this way lends itself to greater portability among domains and to the
possibility of easier acquisition of new patterns.

The advantages of the FASTUS system are as follows:
\begin{itemize}

\item  	It is conceptually simple.  It is a set of cascaded finite-state 
	automata.

\item 	It is effective.  It has been among the leaders in recent 
	evaluations.

\item	It has very fast run time.  

\item   In part because of the fast run time, it has a very fast 
	development time.  This is also true because the system provides
	a direct link between the texts being analyzed and the data
	being extracted.

\end{itemize}

FASTUS is not a text understanding system.  It is an information
extraction system.  But for information extraction tasks, it is perhaps
the most convenient and most effective system that has been developed.

One of the lessons to be learned from our FASTUS experience is that
many information extraction tasks are much easier than anyone ever
thought.  Although the full linguistic complexity of the texts is often
very high, with long sentences and interesting discourse structure
problems, the relative simplicity of the information-extraction task
allows much of this linguistic complexity to be bypassed---indeed much
more than we had originally believed was possible.  The key to the whole
problem, as we see it from our FASTUS experience, is to do exactly the
right amount of syntax, so that pragmatics can take over its share of
the load.  For many information extraction tasks, we think FASTUS
displays exactly the right mixture.

	While FASTUS is an elegant achievement, the whole host of
linguistic problems that were bypassed are still out there, and will
have to be addressed eventually for more complex tasks, and to achieve
higher performance on simple tasks.  We have shown one can go a long way
with simple techniques.  But the hard problems cannot be ignored
forever, and scientific progress requires that they be addressed.

\section*{Acknowledgments}

The FASTUS system was originally built under SRI internal research and
development funding.  Specific applications and improvements have been
funded by the (Defense) Advanced Research Projects Agency under Office
of Naval Research contract N00014-90-C-0220, Office of Research and
Development contract 94-F157700-000, and Naval Command, Control and
Ocean Surveilliance Center contract N66001-94-C-6044, and by the US Army
Topographic Engineering Center under contract no. DACA76-93-L-0019.
Specific developments have also been funded by commercial contracts.

\end{document}